\documentclass{neus2025}

\title[Foundation Models for Certifiable Logistics]{Foundation Models for Logistics Operations: \\ Toward Certifiable, Conversational Interfaces}
\usepackage{times}
\usepackage{algorithm}
\usepackage{algpseudocode}
\usepackage{subcaption}
\usepackage{newfloat}
\usepackage{adjustbox}
\usepackage{listings}
\usepackage[table]{xcolor}
\usepackage{tikz}
\usepackage{mathtools} % for \coloneqq
\usepackage{amsfonts} % 
\usepackage{wrapfig}
\usepackage{subcaption}
\usepackage{booktabs}

\usepackage[most]{tcolorbox}
\usepackage{parskip} % For spacing between paragraphs
\usetikzlibrary{arrows.meta, positioning, shapes.multipart, matrix, fit}

\usetikzlibrary{arrows, automata, positioning}
\tikzset{auto, >=stealth}
\tikzset{every edge/.append style={shorten >= 1pt}}
\tikzset{
    main node/.style={circle,draw,minimum size=1cm,inner sep=0pt},
}

\DeclareCaptionStyle{ruled}{labelfont=normalfont,labelsep=colon,strut=off} % DO NOT CHANGE THIS
\floatstyle{ruled}

\usepackage{listings}
\definecolor{codegreen}{rgb}{0,0.6,0}
\definecolor{codegray}{rgb}{0.5,0.5,0.5}
\definecolor{codepurple}{rgb}{0.58,0,0.82}
\definecolor{backcolour}{rgb}{0.95,0.95,0.92}

\lstdefinestyle{code}{
    backgroundcolor=\color{backcolour},   
    commentstyle=\color{codegreen},
    keywordstyle=\color{magenta},
    numberstyle=\tiny\color{codegray},
    stringstyle=\color{codepurple},
    basicstyle=\ttfamily\footnotesize,
    breakatwhitespace=false,         
    breaklines=true,                 
    captionpos=b,                    
    keepspaces=true,                 
    numbers=left,                    
    numbersep=5pt,                  
    showspaces=false,                
    showstringspaces=false,
    showtabs=false,                  
    tabsize=2,
    moredelim=**[is][\color{blue}]{@}{@}
}

\lstdefinestyle{base}{
  language=tcl,
  emptylines=1,
  breaklines=true,
  basicstyle=\ttfamily\color{blue},
  moredelim=**[is][\color{purple}]{@}{@},
}

\usepackage{graphicx, array, blindtext}
\usepackage{dblfloatfix}

\usepackage{colortbl}

\author{%
 \Name{Yunhao Yang$^{1,2}$\footnote{This work was initiated during the internship of Yunhao Yang at the Neurosymbolic Intelligence.}, Neel P. Bhatt$^{1,2}$, Christian Ellis$^{1,2}$, Samuel Li$^{2}$, Alvaro Velasquez$^{1,3}$, \\ Zhangyang Wang$^{2}$, Ufuk Topcu$^{1,2}$ }\\
 \addr $^1$ Neurosymbolic Intelligence, $^2$ The University of Texas at Austin, $^3$ University of Colorado Boulder
}

\begin{document}

\maketitle

\begin{abstract}
\noindent %
% Logistics operators, from battlefield coordinators rerouting airlifts ahead of a storm to warehouse managers juggling late trucks, often face life‑critical decisions that require both domain expertise and continuous replanning. 
Logistics operators -- from battlefield coordinators re-routing airlifts ahead of a storm to warehouse managers juggling late trucks -- need to make mission‑critical decisions. 
Prevailing methods for logistics planning such as integer programming yield plans that satisfy user-defined logical constraints, assuming an idealized mathematical model of the environment. On the other hand, foundation models lower the intermediate processing barrier by translating natural-language user utterances into executable plans, yet they remain prone to misinterpretations and hallucinations that jeopardize safety and cost.
We introduce a \emph{Vision–Language Logistics} (VLL) agent, built on a neurosymbolic framework that pairs the accessibility of natural‑language dialogue with \textit{verifiable guarantees on user-objective interpretation}.
The agent interprets user requests and converts them into structured planning specifications, quantifies the uncertainty of the interpretation, and invokes an interactive clarification loop when the uncertainty exceeds an adaptive threshold.
Drawing on a lightweight airlift logistics planning use case as an illustrative case study, we highlight a practical path toward \emph{certifiable and user-aligned} decision-making for complex logistics.
Our lightweight model, fine‑tuned on just 100 training samples, surpasses the zero‑shot performance of 20x larger models in logistic planning tasks while cutting inference latency by nearly 50\%. Our code and video demonstrations are available at \href{https://github.com/samuelLi05/airplane_simulator}{\textcolor{blue}{here}}.
\end{abstract}


\section{Introduction}
\label{sec:introduction}

% Logistics technology is stuck in the past
% + The generated plan is guaranteed to satisfy constraints
% - Assumes perfect situational awareness
% - Slow (this results took 50+ hours)
% - Incapable of rapid replanning
% - Incapable of ingesting multiple modalities of data

% LLMs are not fit for logistics
% + The generated plan can be interacted with and recomputed quickly
% + The LLM can ingest multiple modalities of data
% + The LLM has a context window of memory that makes result take into account historical information and on-the-fly natural text inputs from end users
% - The generated plan does not satisfy constraints

% Change cognition to reasoning and connect to LRM-like use of verifiers plus RL to fine-tune?

\textbf{Toward Conversational, Certifiable Logistics. }%
From humanitarian air‑lifts that must beat incoming storms to e‑commerce warehousing that re-optimizes routes every minute, modern logistics is increasingly a race against time, uncertainty, and domain complexity \citep{ozdamar2004emergency, ghiani2004introduction}. Operators juggle heterogeneous data—live satellite imagery, weather forecasts, inventory databases, and a lattice of hard constraints on safety, cost, and regulation. Today, interacting with these systems demands fluency in specialized planning languages and manual cross‑checks, leaving non‑experts on the sidelines and slowing even seasoned professionals when crises strike \citep{nicoletti2024green, nicoletti2025impact}. This challenge motivates growing interest in more accessible, conversational interfaces for logistics planning.

% Across AI planning and logistics research, there is growing discussion of how multimodal foundation models might serve as accessible co-pilots for complex decision-making. 
We develop \emph{\textbf{Vision–Language-Logistics (VLL)  agents}}: multimodal AI co‑pilots that can (i) understand rich, free‑form instructions; (ii) ground those instructions in real‑time perceptual data; (iii) synthesize and visualize executable plans through tight neurosymbolic integration; and (iv) \emph{prove to the user}, via explicit uncertainty signals and formal checks, that the inferred goals (i.e., user objectives) align with the user’s true intent. VLL agents build on parallel advances in vision–language modeling, uncertainty estimation, and neurosymbolic verification, situating these techniques within the emerging landscape of conversational, certifiable logistics systems.

\begin{figure*}[t]
    \centering
    \includegraphics[width=0.9\textwidth]{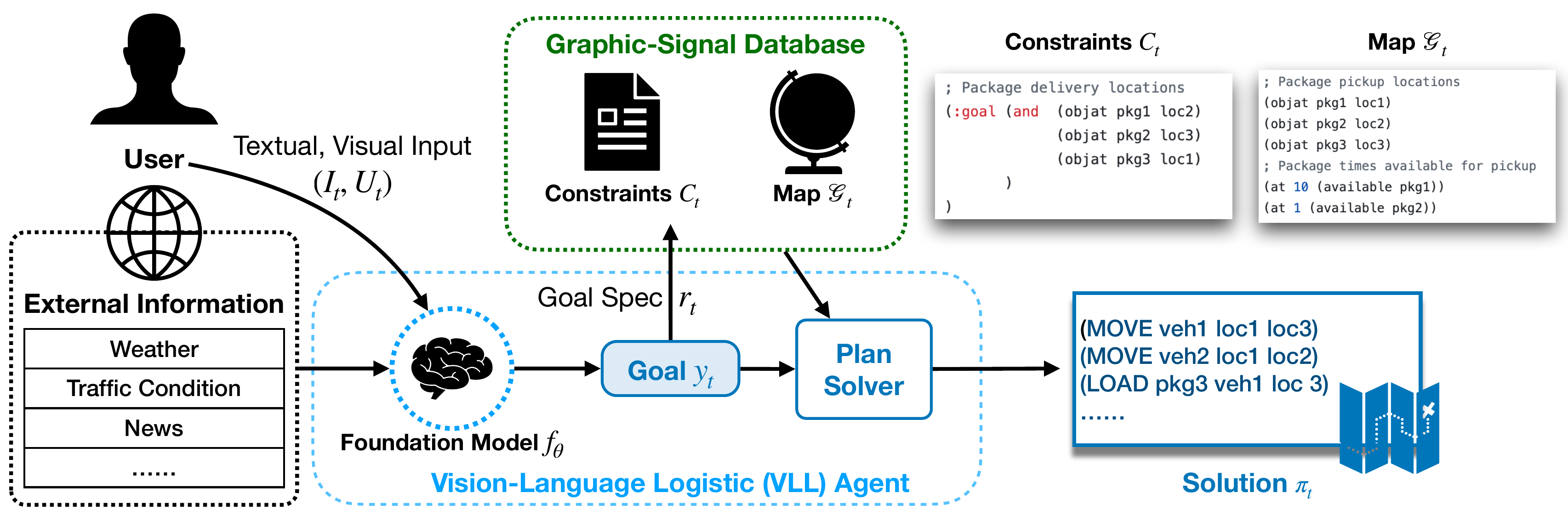}
    \caption{Overview of an VLL agent. Language and visual inputs are converted into structured goals, filtered through an uncertainty‑aware verifier, and dispatched to symbolic planners.}
    \label{fig:vllm_architecture}
\end{figure*}

% \vspace{0.5em}\noindent
% \textbf{High-level framework.} %
A VLL agent is a closed perception–reasoning–action loop. At the perception stage, vision models extract salient entities (e.g., damaged runways or congested highways) that ground the referents in a natural‑language command. A language model then translates the grounded request into a structured goal specification (e.g., PDDL or SQL) that downstream planners can solve. Finally, a symbolic verifier or constraint solver confirms feasibility and compliance, returning either a certified plan or a targeted clarification query. This architecture, depicted in Figure~\ref{fig:vllm_architecture}, enables conversational replanning at the cadence of operational tempo, i.e., users can issue follow‑up directives, receive intermediate analyses, and iterate until the plan is both satisfactory and formally sound. In short, VLL agents promise \emph{accessibility without sacrificing reliability}.

\vspace{0.5em}\noindent
\textbf{A Key Component: Uncertainty‑Aware Intent–Verification.} %
Among the many moving parts of a VLL agent, we concentrate on a critical yet often underserved component: \emph{measuring and reducing the uncertainty in goal classification}.  
Standard LLM/VLM pipelines silently commit to their first parse of a request; if that parse is wrong, downstream optimizers will dutifully compute an optimal solution to the \emph{wrong problem}. The VLL agent augments its VLM backbone with a probabilistic \emph{field‑level uncertainty estimator}. Whenever the estimated uncertainty value in any essential slot—e.g., destination airbase or delivery deadline—falls below an adaptive threshold, the agent pauses, asks a pointed follow‑up question, and only proceeds once uncertainty is resolved.
% The interaction is, therefore, \emph{proactive rather than reactive}: potential misalignments are surfaced before plans are generated, saving computation and, more importantly, preventing costly execution errors.

Technically, the verification loop maps user goals to a learned latent space, where different goal types form distinct clusters. Building on the latent space, we introduce a \emph{Probabilistic Guarantee}: a theoretical lower bound of correct goal classification measured via the distance from a new input’s latent vector to the nearest cluster centroid. Samples with high guarantee indicate low uncertainty. These low-uncertainty traces feed a contrastive self-training loop that steadily sharpens the VLL agent's goal classification ability, allowing it (using the GPT-4o-mini backbone) to \emph{outperform a 20x larger} GPT-4.1 model on goal-match accuracy while halving inference latency.

\vspace{0.5em}\noindent
\paragraph{Contributions.} %
We (i) introduce the VLL agents as a neurosymbolic framework for conversational logistics planning, (ii) propose an uncertainty-aware intent–verification loop that maps multimodal inputs into a learned latent space and provides a probabilistic guarantee on correct goal interpretation, enabling proactive clarification before planning, and (iii) empirically show how this guarantee can guide rapid refinement through fine-tuning and prompt optimization. Experiments show that lightweight models trained on a limited number of high-guarantee samples achieve higher goal classification accuracy.

\section{Related Work}

% Recent advances in AI are reshaping logistics, from natural language interfaces that lower barriers to planning, to uncertainty-aware methods that flag and resolve ambiguity, to neurosymbolic approaches that bring verification into the loop. 
% Logistics applications such as airlift scheduling and warehouse routing provide concrete testbeds where these trends converge.

\paragraph{Practical applications and benchmarks in logistics.}
Research in logistics and transportation supports practical demands and provides benchmarks for evaluation. Classical formulations, for example dynamic pickup-and-delivery \cite{berbeglia2010dynamic} and airlift planning under operational constraints \cite{bertsimas2019airlift} -- establish the cost, capacity, and timing structures that real systems must comply with. Recently, the Airlift Challenge frames cargo delivery as a community benchmark with dynamic conditions \cite{delanovic2024airlift}. 
Against this backdrop, LLM/VLM-based systems are increasingly explored as front-ends that translate informal directives into structured goals. Yet, most existing demonstrations treat models as black-box solvers \citep{nicoletti2025impact, nicoletti2024green}. We emphasizes transparent goal interpretation and compatibility with symbolic verifiers, aiming to make conversational interfaces viable for logistics operations.

% LLMs have been recently explored for planning and decision-making in logistics and control tasks \cite{berbeglia2010dynamic, bertsimas2019airlift, delanovic2024airlift}. However, existing approaches often treat LLMs as black-box solvers, focusing on task execution rather than alignment and verification. Our work departs from this line by focusing on goal understanding and structured reformulation, while incorporating uncertainty-aware mechanisms to support human-in-the-loop verification.

\paragraph{Natural language interfaces for planning.}
Recent advances in LLMs have significantly improved the ability to convert natural language into formal planning representations. Prior work has explored using LLMs to translate high-level goals into formats such as PDDL, JSON, or action graphs, enabling non-expert users to interact with complex systems \citep{Song2022LLMPlannerFG, brohan2023can, shah2022robotic, huang2022inner, hao2024planning}. 
% Specifically, LLM-Planner demonstrates few-shot grounded planning by mapping language to action graphs that interact with the environment \cite{Song2022LLMPlannerFG}. SayCan grounds language in robot affordances to select skills consistent with what the robot can do \cite{brohan2023can}. LM-Nav combines scene understanding with language guidance for navigation over long horizons \cite{shah2022robotic}. Beyond skill selection, LLM+P explicitly composes LLMs with optimal planners to keep search optimality while benefiting from language priors \cite{Liu2023LLMPEL}.
Collectively, these works lower barriers for non-experts and hence improve accessibility of the planner. On the downside, they assume that the LLMs interpret user objectives correctly and lack verification mechanisms to ensure plan correctness or goal alignment.

\paragraph{Uncertainty estimation in LLMs.}
Uncertainty estimation in LLMs has gained increasing attention as a mechanism to detect hallucinations, flag low-confidence predictions, or guide interaction. Techniques such as token-level entropy, Monte Carlo dropout, and calibration-based methods have been applied to identify unreliable outputs \citep{gupta2024language, milanes2021monte, yang2024uncertainty, Bhatt2025KnowWhere}. 
% More recent work integrates uncertainty into the model's training loop to steer learning or adjust user interaction strategies \citep{Bhatt2025KnowWhere}. 
Together, these works motivate proactive interaction loops---surfacing ambiguity before committing to a plan---but most stop short of offering guarantees on intent correctness in logistics contexts.
% Our approach builds on this idea by using uncertainty estimates not only to flag low-confidence outputs but also to initiate clarification loops with the user.

\paragraph{Neurosymbolic and verification-informed planning.}
Recent works combines neural models with symbolic verification or formal reasoning systems for robotics and planning tasks, offering correctness guarantees \citep{chen2025learning, Yang2024, li2024formal, Yang2024AAMAS}. 
% Recent work uses decision-boundary distances to provide probabilistic guarantees on specification compliance \citep{chen2025learning}. These directions suggest a path toward certifiable interfaces: use symbolic structure to check or shape the outputs of foundation models. 
The VLL agent extends the neurosymbolic and verification ideas to the problem of natural language goal interpretation in logistics, using uncertainty-aware learning to balance accessibility and verifiability.
\section{Problem Formulation}
\label{sec:vll_agent}

\begin{figure*}[t]
    \centering
    \includegraphics[width=0.95\textwidth]{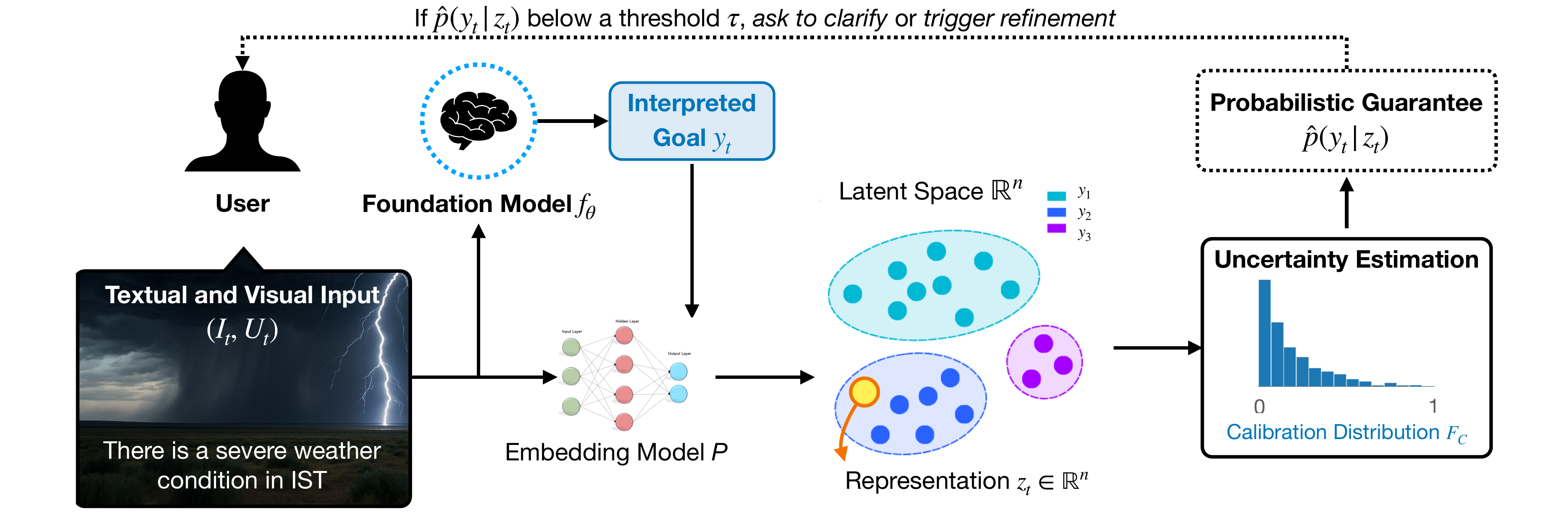}
    \caption{Overview of the uncertainty-aware intent-verification loop.
    % The agent takes multimodal inputs—textual instructions and visual data—and embeds them into a learned latent space where intent types form distinct clusters. It computes a probabilistic guarantee from the distance to the nearest cluster centroid, and, if the guarantee falls below a threshold, proactively issues a clarification query before formalizing the goal to ensure downstream planners work with the correct intent.
    }
    \label{fig: pipeline}
\end{figure*}

We develop a \emph{Vision--Language Logistics (VLL) agent}: a neurosymbolic system that transforms free-form, multimodal user instructions into certified, executable logistics decisions. The key challenge is that user intent is often implicit or ambiguous, while downstream planners and solvers require precise, structured specifications. We formalize two coupled problems: \textbf{(P1)} end-to-end VLL agent construction with a three-stage architecture, and \textbf{(P2)} uncertainty-aware intent verification concentrated in the reasoning stage.

\paragraph{Notation and Model}
Let $t$ be an index. The user provides a multimodal prompt
\[
x_t \triangleq (I_t, U_t) \in \mathcal{X},
\]
where $I_t$ is a natural-language instruction and $U_t$ is a visual context (e.g., satellite imagery, camera feeds, annotated maps).
Let $\mathcal{R}$ denote a space of structured planning representations (e.g., PDDL instances, SQL queries, or symbolic task graphs), and let $\mathcal{Y} = \{1,\dots,K\}$ denote a finite set of user goals (e.g., \emph{query database}, \emph{update database}, \emph{solve a routing problem}). We denote by $y_t^\star \in \mathcal{Y}$ the user’s true goal.

\paragraph{Design an End-to-End VLL Agent with Three Stages.}

A VLL agent is a closed-loop system composed of a perception module, a foundation model, and downstream solver.

A \textbf{perception} module processes visual inputs $U_t$ to extract salient entities, attributes, and situational cues, producing a grounded state abstraction $s_t \in \mathcal{S}.$
This stage resolves linguistic referents such as ``the nearest operational airbase'' or ``avoid flooded regions.''

A foundation model $f_{\theta}$ \textbf{reasons} the multimodal inputs and maps them into a structured specification $r_t$ and a goal label $y_t$:
\[
(r_t, y_t) = f_{\theta}(I_t, s_t) \in \mathcal{R} \times \mathcal{Y}.
\]

A downstream solver produces a candidate plan $\pi_t = \mathrm{Solve}(r_t)$, which is checked against domain constraints $\mathcal{C}_t = \mathcal{C} \cup r_t$ ($\mathcal{C}$ is a set of universal constraints like safety rules, regulations, resource limits) via a symbolic verifier. If verification fails, the agent returns diagnostic feedback and updates the representation or user prompt.

\paragraph{Develop an Uncertainty-Aware Intent Verification Loop.}

Downstream planners assume that the inferred goal $y_t$ correctly reflects the user’s intent. However, misclassification at the reasoning stage can lead to optimal solutions for the \emph{wrong problem}. We therefore require the reasoning stage to be uncertainty-aware and certifiable.

Given input $x_t$, the reasoning module outputs not only a predicted goal $y_t$, but also a probabilistic guarantee
\[
\hat{p}_t \triangleq \hat{p}(y_t \mid x_t),
\]
which lower-bounds the probability that $y_t$ equals the true goal $y_t^\star$:
\[
\hat{p}(y_t \mid x_t) \le \Pr(y_t = y_t^\star \mid x_t).
\]

% \begin{figure}[t]
%     \centering
%     % \vspace{-10pt}
%     \centering
%     \input{figure/vll-example}
%     \caption{Examples demonstrating the VLL agent generating planning solutions or updating the database based on user queries.}
%     \label{fig:qual_examples}
% \end{figure}

% \begin{wrapfigure}{r}{0.35\textwidth}
%     \centering
%     % \vspace{-10pt}
%     \centering
%     \input{figure/vll-example}
%     \caption{\color{blue} UPDATE FIG Examples demonstrating the VLL agent generating planning solutions or updating the database based on user queries.}
%     \label{fig:qual_examples}
%     \vspace{-25pt}
% \end{wrapfigure}

\paragraph{Uncertainty-Guided Refinement.}
Develop a reasoning-stage mechanism that computes $\hat{p}(y_t \mid x_t)$, and uses this guarantee to guide both interactive clarification and uncertainty-filtered fine-tuning of the foundation model $f_{\theta}$.
Formally, given a stream of interaction data, we aim to maximize correct goal inference conditioned on acceptance:
\[
\max_{\theta}\;
\Pr\!\left(f_{\theta}(x)=y^\star \;\middle|\; \hat{p}(f_{\theta}(x)\mid x)\ge\tau\right),
\]
while controlling the clarification frequency induced by $\tau$.

\section{Vision-Language Logistic Agent}

We design the VLL agent as a modular, closed-loop pipeline that integrates multimodal perception, symbolic planning, and formal verification.

At each interaction round $t$, the agent receives visual+textual inputs $(I_t, U_t)$. These inputs are processed to a foundation model $f_{\theta}$ that translates the multimodal input into a structured \emph{goal specification} $r_t$ expressed in the PDDL \citep{pddl}, which serves as the interface to downstream planners. Then, given a PDDL goal specification $r_t$, an off-the-shelf PDDL solver is invoked to synthesize a plan $\pi_t$. We present the architecture in Figure \ref{fig:vllm_architecture}.
% To ensure correctness beyond plan feasibility, the resulting plan is checked against domain-specific constraints expressed in temporal logic (e.g., safety, resource, or regulatory requirements). A model checker verifies whether the plan satisfies these temporal specifications, returning either a certificate of compliance or a counterexample indicating violation.

Crucially, the reasoning stage does not commit to a PDDL goal unless the inferred user intent (i.e., goal) has been verified with sufficient confidence. We propose an uncertainty-aware intent-verification loop that acts as a gatekeeper between multimodal interpretation and symbolic planning, ensuring that formal verification is applied only to goals that are certified to reflect user intent. We present the online interface of the VLL agent with more demonstrations in the \href{https://github.com/samuelLi05/airplane_simulator}{\textcolor{blue}{code repository}}.

\subsection{Uncertainty-Aware Intent-Verification}
\label{sec:method}

We develop an uncertainty-aware intent-verification loop that (i) infers the user goal from multimodal inputs, (ii) computes a calibrated probabilistic guarantee on intent correctness, and (iii) triggers interactive clarification when the guarantee is insufficient. The goal is to ensure that downstream planning operates only on objectives that are both well-formed and certified to reflect user intent. We present this loop in Figure \ref{fig: pipeline}. 

We focus exclusively on the \emph{reasoning stage} of the VLL agent. Perception and symbolic verification are treated as black-box components that provide grounded inputs and constraint checks, respectively. Our contribution is an intent-verification loop that sits between perception and planning and provides formal guarantees on goal classification.

\paragraph{User Goal Classification.}

Given a multimodal input $x_t = (I_t, U_t)$, the VLL agent uses a foundation model $f_{\theta}$ to infer the user’s goal.
Concretely, the model treats intent inference as a classification problem over a set of goals $\mathcal{Y}$. The foundation model outputs $y_t = f_{\theta}(x_t),$ where $y_t$ is the predicted label corresponding to the user’s high-level goal (e.g., information retrieval, database update, or logistics planning).

\begin{figure}[t]
    \centering
    \includegraphics[width=0.3\linewidth]{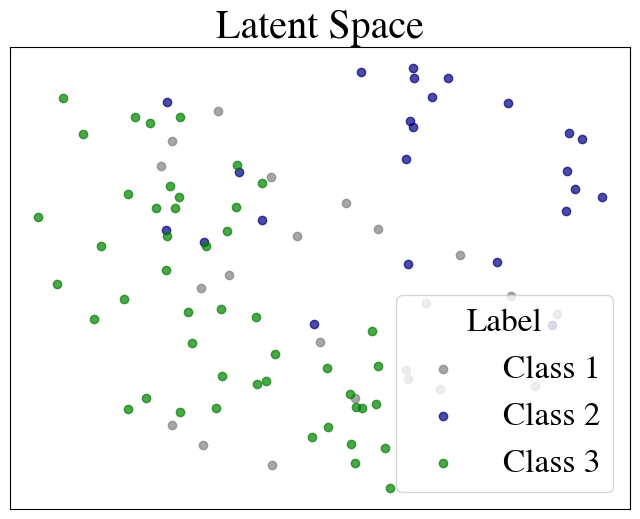}
    \includegraphics[width=0.3\linewidth]{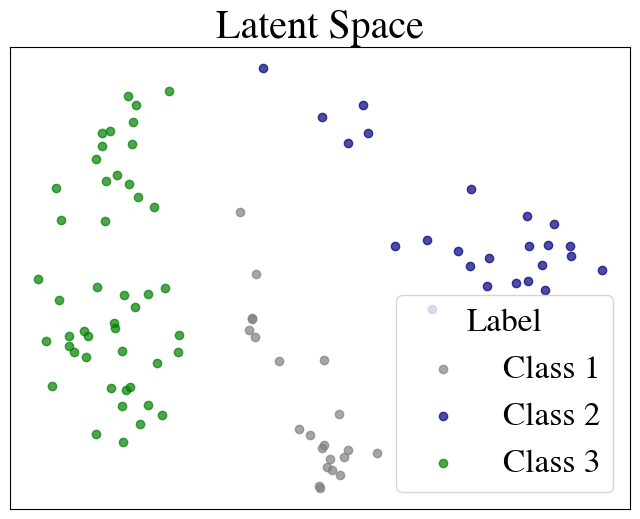}
    \includegraphics[width=0.3\linewidth]{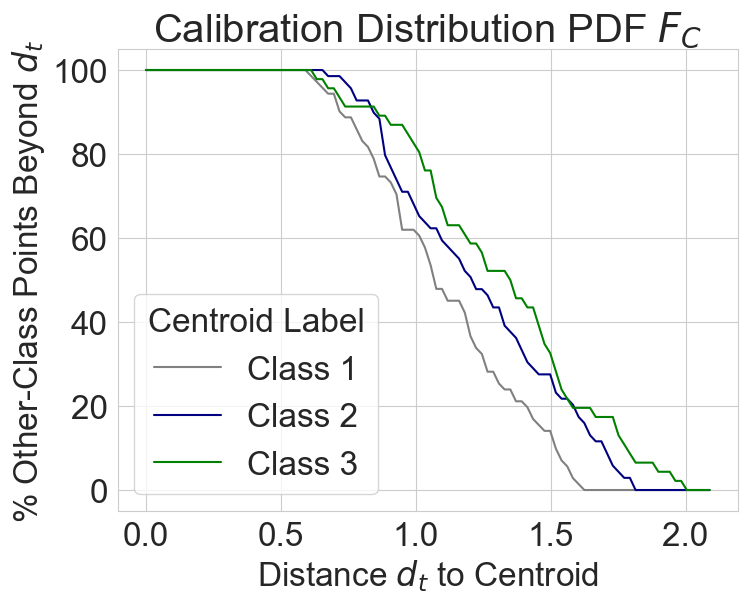}
    \caption{The left plot is a latent space before learning. The middle plot is a learned space produced from a fine-tuned model, where three classes of user goals are separated. The right plot shows the calibration distributions of the three classes estimated via the learned latent space.}
    \vspace{-10pt}
    \label{fig: demo}
\end{figure}

\paragraph{Latent Space Learning.}
We train an embedding model $P$ to map multimodal inputs and their predicted goals into an $n$-dimensional latent space $\mathbb{R}^n$ in which distinct user goals form well-separated clusters. Using 400 manually labeled examples, we optimize $P$ with a supervised contrastive loss. 
Let $z_i = P(y_i, I_i, U_i) \in \mathbb{R}^n$ denote the latent representation with label $y^*_i$. For a minibatch $\mathcal{B}$ and positive set $\mathcal{P}(i)=\{j\in\mathcal{B}\setminus\{i\}:y_j=y_i\}$, we minimize
\[
\mathcal{L}_{\mathrm{con}}
=
\sum_{i \in \mathcal{B}}
\frac{-1}{|\mathcal{P}(i)|}
\sum_{j \in \mathcal{P}(i)}
\log
\frac{
\exp\!\left( z_i^\top z_j / \tau \right)
}{
\sum_{k \in \mathcal{B} \setminus \{i\}}
\exp\!\left( z_i^\top z_k / \tau \right)
},
\]
where $\tau>0$ is a temperature parameter. Figure~\ref{fig: demo} illustrates the learned latent space for a three-class goal classification task (see \textbf{Empirical Analysis}).

\paragraph{Guarantee on Classification Correctness.}
Building on the clustered latent space, we formalize how distances in the latent space are calibrated to a likelihood of classification correctness.

We define the \emph{probabilistic guarantee} as a lower bound on the probability that the predicted goal $y_t$ matches the user’s true goal $y_t^*$ given the latent representation $z_t = P(y_t, I_t, U_t) \in \mathbb{R}^n$:
\[
\hat{p}(y_t \mid z_t) = \Pr\big[ y_t = y_t^{\ast} \,\big|\, z_t \big].
\]
To estimate $\hat{p}(y_t \mid z_t)$, we use a small calibration set of 200 manually labeled examples to construct a calibration distribution with probability density function (PDF) $F_C$. $F_C$ maps the distance $d_t = || z_t - c^* ||_2$ from $z_t$ to its nearest cluster centroid $c^*$ to the probability that a sample from another class lies beyond that distance
\begin{center}
    $\Pr\big[ || z_i - c^* ||_2 > d_t \,\big|\, y_i \ne y_c^* \big]$, 
\end{center}
$z_i$ denotes any latent vector and $y_c^*$ denotes the label of the cluster where $c^*$ lies in. Figure \ref{fig: demo} shows an example of $F_C$ obtained from the latent representations.

Given a new input, we compute its distance $d_t$ to the nearest centroid and evaluate
\begin{equation}
    \hat{p}(y_t \mid z_t) = 1 - \frac{\big(1 - F_C(d_t)\big)\cdot \Pr\big[y_i \ne y_c^*\big]}{\Pr\big[ || z_i - c^* ||_2 \le d_t \big]}.
\end{equation}
$\Pr\big[y_i \ne y_c^*\big]$ is the number of other-class samples over the total number of samples, and $\Pr\big[ || z_i - c^* ||_2 \le d_t \big]$ is the percentage of samples whose distance to $c^*$ is within $d_t$.
Higher values of $\hat{p}(y_t \mid z_t)$ indicate greater confidence in the predicted goal. We present the theorems and proofs for the probabilistic guarantee in the Appendix \ref{app: theory}.

\paragraph{Proactive Clarification.}
When the probabilistic guarantee $\hat{p}_t = \hat{p}(y_t \mid z_t)$ falls below a predefined threshold $\tau$, the VLL agent proactively enters a clarification mode before passing the goal to the planner. In this mode, the agent implements a clarification loop governed by a threshold $\tau \in (0,1)$:
\begin{itemize}
    \item If $\hat{p}_t \ge \tau$, the inferred goal $y_t$ is accepted and passed to the planner.
    \item If $\hat{p}_t < \tau$, the agent issues a targeted clarification query to the user (e.g., ``Do you mean to minimize cost or time?'' or ``Should weather conditions be considered?'').
\end{itemize}

This loop terminates at a stopping time $T$ such that
$
\hat{p}(y_T \mid x_T) \ge \tau,
$
ensuring that planning proceeds only after sufficient confidence in the intent correctness.

% \begin{figure*}[t]
%     \centering
%     \input{figure/vll-example}
%     \caption{Examples demonstrating our VLL agent.}
%     \label{fig:qual_examples}
% \end{figure*}

\subsection{Uncertainty-Guided Refinement}
\label{sec: refine}
We then develop two refinement methods that leverage the calibrated uncertainty signal, i.e., probabilistic guarantee, to improve the VLL agent’s goal classification. 

\paragraph{DPO with Guarantee Ranking.}
The first method uses the probabilistic guarantee to induce preference rankings for Direct Preference Optimization (DPO) \citep{dpo}. Given a user input $x_t$ and a set of candidate interpretations $\{y_t^{(k)}\}_{k=1}^K$ produced by the foundation model, we compute their guarantees
$\hat{p}_t^{(k)} \triangleq \hat{p}\!\left(y_t^{(k)} \mid x_t\right).$
For any pair $(i,j)$ such that $\hat{p}_t^{(i)} > \widehat{p}_t^{(j)}$, we rank a pair $y_t^{(i)} \succ y_t^{(j)}$. These ranked pairs are used to fine-tune the model via DPO by maximizing the likelihood ratio between preferred and dispreferred goals.

\paragraph{TextGrad with Online Uncertainty Feedback.}
In addition to model-parameter fine-tuning, we employ TextGrad \citep{yuksekgonul2024textgrad} to optimize the system prompt using online interaction feedback. We convert the probabilistic guarantee into auto-generated textual feedback (e.g., ``the inferred goal is correct with approximately 75\% confidence''), which serves as a differentiable supervision signal for prompt optimization.

\paragraph{Generality Beyond Logistics.} Although demonstrated in a logistics setting, the uncertainty-aware intent-verification loop is domain-agnostic and can be applied to any multimodal system requiring reliable goal interpretation, such as robotic execution, autonomous driving, database querying, and decision-support systems.
\section{Empirical Analysis}
\label{sec:experiment}

We evaluate the intent verification pipeline through quantitative improvements in model performance, demonstrating that proactive clarification and uncertainty-guided refinement significantly improve the model's ability to infer user goals in the context of logistics planning.

\begin{figure*}[t]
    \centering
    \includegraphics[width=\textwidth]{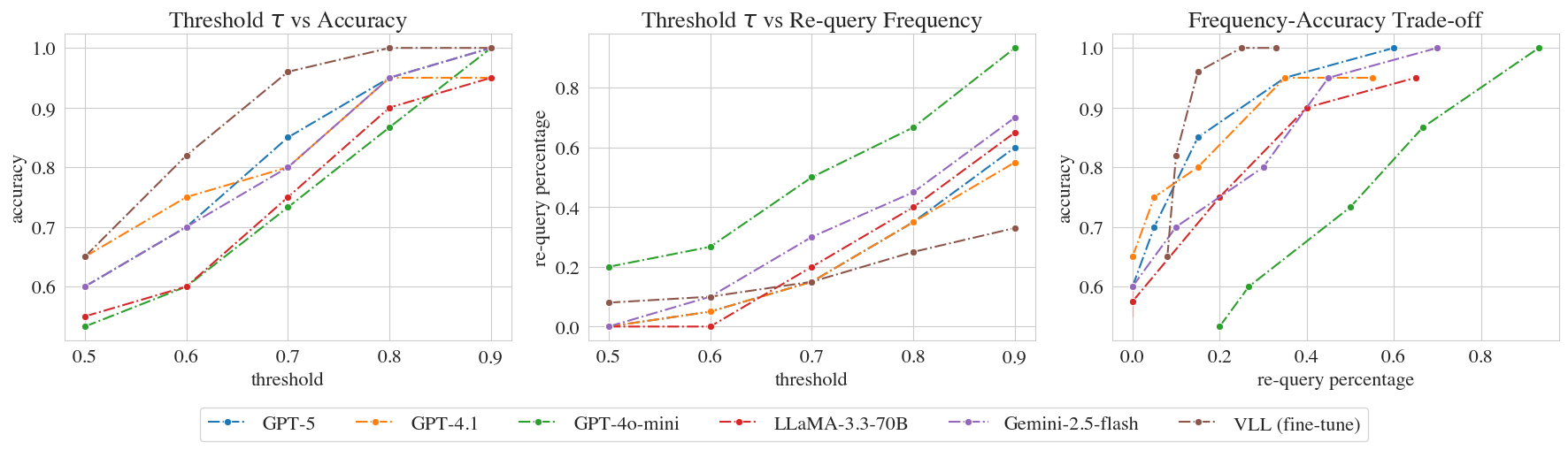}
    \caption{
    Comparison on goal classification accuracies and re-query frequencies. The fine-tuned VLL achieves better performance than all other baselines with a lower frequency of re-query, showing the effectiveness of uncertainty-guided fine-tuning in aligning model behavior with user goals.
    \vspace{-10pt}
    }
    \label{fig:quant_results}
\end{figure*}

\paragraph{Experimental Setting.}
We evaluate the VLL agent on a lightweight airlift logistics planning domain designed to stress \emph{goal interpretation under uncertainty}, rather than downstream optimization quality.
The learning task is to infer the user’s high-level goal
\(y_t \in \mathcal{Y}\),
where
\(
\mathcal{Y} = \{\text{Query}, \text{Update}, \text{Plan}\}
\)
corresponds to (i) extracting information from a logistics database, (ii) updating the database, or (iii) solving a logistics planning problem.
These categories capture the dominant failure modes in conversational logistics systems, where an incorrect goal classification leads to valid but operationally irrelevant plans.

We construct a dataset of 400 manually labeled interaction traces to learn the latent space, and an additional 200 labeled samples to estimate the calibration distribution for the probabilistic guarantee. For refinement, we follow Section \ref{sec: refine} to create 100 samples for each refinement approach. During evaluation, we use 200 testing samples that are independent to the training and calibration data.

\begin{figure*}[t]
    \centering
    \includegraphics[width=0.9\textwidth]{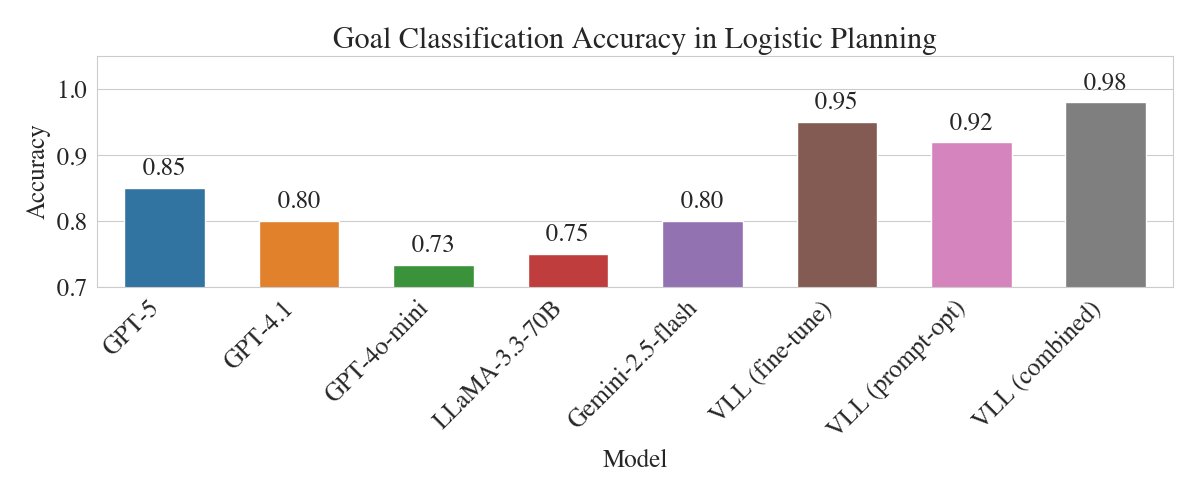}
    \vspace{-15pt}
    \caption{
    Comparison between the baseline foundation models and our refined VLL agents. ``VLL (combined)'' refers to the fine-tuned GPT-4o-mini backbone with optimized prompts. The refined VLL with 4o-mini backbone outperforms the 20x larger models such as GPT-5.
    \vspace{-5pt}
    }
    \label{fig:quant_results_ext}
\end{figure*}

\paragraph{Baselines.}
We compare the VLL agent against a set of foundation-model baselines summarized in Figure \ref{fig:quant_results_ext}.
For all baselines, we \emph{directly query} the model for \emph{goal classification} without any verification or clarification loop.
Specifically, given a multimodal input \(x_t = (I_t, U_t)\), the model is prompted with an instruction of the form:
\emph{``Classify the user goal based on the given inputs into one of the following categories: Query, Update, or Plan.''}
We extract the predicted goal \(y_t\) from the model's output.

We consider 6 baseline models as presented in Figure \ref{fig:quant_results_ext}, all evaluated in a zero-shot setting using the same prompt template.

\paragraph{Evaluation Metric.}
We evaluate the impact of the proposed uncertainty-guided fine-tuning on the VLL agent’s foundation model $f_\theta$ (GPT-4o-mini backbone) and use three metrics:

\noindent
\textit{Threshold} is a scalar cut-off value $\tau$ to the probabilistic guarantee $\hat{p}_t = \hat{p}(y_t \mid z_t)$. Only samples with scores above $\tau$ are retained for accuracy evaluation.

\noindent
\textit{Accuracy} is the proportion of data samples whose predicted intent $y_t$ matches the ground-truth intent $y_t^{\ast}$, computed over the subset of samples that pass the threshold filter:
\vspace{-6pt}
\[
\text{Accuracy} = \#\{\, y_t = y_t^{\ast} \;\wedge\; \hat{p}(y_t \mid z_t) \ge \tau \,\} \ / \ \#\{\, \hat{p}(y_t \mid z_t) \ge \tau \,\}.
\vspace{-6pt}
\]
\noindent
\textit{Re-query Frequency} is the proportion of evaluation samples for which the uncertainty signal falls below the threshold $\tau$, triggering a clarification query to the user.
Lower values indicate fewer interruptions for clarification, while higher values suggest more frequent user interaction to resolve intent ambiguity.

\paragraph{Results and Analysis.}
Figures \ref{fig:quant_results} and \ref{fig:quant_results_ext} present the quantitative results of uncertainty-aware refinement on goal classification.
Figure \ref{fig:quant_results} illustrates the trade-off between goal classification accuracy and re-query frequency as the acceptance threshold \(\tau\) varies.
Across all thresholds, VLL consistently achieve higher accuracy for a fixed clarification rate compared to zero-shot foundation models.

Figure \ref{fig:quant_results_ext} compares the VLL agents against increasingly larger foundation-model baselines.
Despite using a significantly smaller backbone, the VLL outperforms \textbf{20x larger models} such as GPT-5 in goal classification accuracy while inducing faster response time.
Notably, the combined refinement strategy (fine-tuning + prompt optimization) yields the strongest overall performance, showing the effectiveness of our uncertainty-guided refinement strategies.

Table \ref{tab: prompt} provides qualitative insight into this improvement by illustrating prompt optimization driven by uncertainty feedback.
In each example, the refined prompt produced by the VLL agent is more explicit and operationally grounded, leading to a higher probabilistic guarantee \( \hat{p}_t \).
Together, these results indicate that uncertainty-aware verification and refinement improve both the correctness of inferred goals and the clarity of model reasoning, which are critical for reliable conversational logistics systems.

\begin{table}[t]
    \centering
    \begin{tabular}{m{0.35\linewidth} m{0.07\linewidth} m{0.35\linewidth} m{0.07\linewidth}}
    Initial Prompt & $\hat{p}_t$ & Optimized Prompt & $\hat{p}_t$ \\
    \hline
    A blocking condition appears in the route between airports 1 and 5 that needs to be shown. & 0.607 & Change the value of `route available' to false for all entries where `origin airport id' equals 1 and `destination airport id' equals 5 & 0.722 \\
    \hline
    Analyze routes from airport 3 are consistent and cross-check inconsistent prices and cargoes. & 0.669 & Retrieve all routes where airport id = 3; compare entries for matching destinations to identify discrepancies in price and cargo type. & 0.738 \\
    \hline
    \end{tabular}
    \caption{Examples of uncertainty-guided prompt optimization.
    \vspace{-10pt}
    }
    \label{tab: prompt}
\end{table}

\paragraph{Ablation Study.}
We conduct an ablation study to assess the necessity of each component in the uncertainty-aware intent-verification and refinement.
All variants share the same foundation model backbone and training data.
We present the classification accuracy and average classification latency per query in Table \ref{tab:ablation}. For each variant, we use the corresponding uncertainty signal to fine-tune the backbone foundation model and present the accuracy and the average inference time. There is no fine-tuning if the signal is not specified. 
We show that removing any single component degrades the classification accuracy, while does not significantly reduce the latency.

\begin{table}[t]
\centering
\small
\begin{tabular}{l l l l c c c}
\toprule
\textbf{System Variant} 
& \textbf{Projector} 
& \textbf{Loss} 
& \textbf{Signal} 
& \textbf{Fine-Tuned} 
& \textbf{Accuracy} 
& \textbf{Latency (s)}\\
\midrule
\textbf{VLL (fine-tune)} 
& Learned 
& Contrastive 
& Guarantee 
& \checkmark
& \textbf{0.96}
& 1.33 \\

FM only (zero-shot) 
& -- 
& -- 
& -- 
&
& 0.73 
& 1.20 \\

Projector only 
& Learned 
& Contrastive 
& -- 
& \checkmark
& 0.65 
& 0.15 \\

No projector $P$ 
& -- 
& -- 
& Softmax 
& \checkmark
& 0.75 
& 1.28 \\

Frozen projector 
& Frozen 
& Contrastive 
& Guarantee 
& \checkmark
& 0.81 
& 1.33 \\

Supervised
& Learned 
& Cross Entropy 
& Guarantee 
& \checkmark
& 0.91 
& 1.34 \\

Softmax 
& Learned 
& Contrastive 
& Softmax 
& \checkmark
& 0.92 
& 1.29 \\

\bottomrule
\end{tabular}
\caption{Ablation study over different projectors, projector training losses, uncertainty signals for ranking DPO data and clarification, and fine-tuned/zero-shot foundation models. Among the variants, \emph{FM only} refers to directly querying the backbone foundation model for classification. \emph{Projector only} is an MLP trained for goal classification using labeled data. \emph{No projector} removes latent space and relies on model confidence alone. \emph{Frozen projector} uses a fixed, untrained latent mapping. \emph{Supervised} replaces contrastive learning with supervised learning. \emph{Softmax} replaces probabilistic guarantee with the softmax confidence of each prediction.
\vspace{-15pt}
}
\label{tab:ablation}
\end{table}
\section{Conclusion}
We introduce a neurosymbolic framework for
conversational logistics that prioritizes \emph{certifiable goal interpretation} over black-box plan generation.
By mapping multimodal user inputs into a learned latent space and computing a calibrated probabilistic guarantee on goal correctness, the VLL agent can proactively clarify ambiguous user requests and guide uncertainty-aware refinement to its backbone foundation model.
Empirical results show that these mechanisms enable smaller foundation models to outperform substantially larger ones in goal classification accuracy while reducing unnecessary clarification, demonstrating that structured uncertainty signals are more critical than raw model scale for reliable conversational logistics.

% \paragraph{Future Directions.}
Future work will extend VLL agents beyond intent verification to end-to-end certifiable logistics pipelines. Potential directions include integrating symbolic proof engines to generate machine-checkable certificates of plan correctness, supporting \emph{real-time grounding} from streaming perceptual inputs, and enabling \emph{cache-guided plan updates} under changing conditions. 
% These extensions move toward conversational logistics systems that are not only flexible and efficient, but also provably aligned with user intent and operational constraints.

\bibliography{reference}

\newpage
\appendix
\onecolumn
\section{Theoretical Support for Probabilistic Guarantee}
\label{app: theory}

We provide theoretical supports for the probabilistic guarantee $\hat{p}(y_t \mid z_t)$ used in our uncertainty-aware intent–verification pipeline. Specifically, we show how distances in the learned latent space, constructed from multimodal embeddings of textual and visual inputs, can be mapped to calibrated confidence estimates on user intent classification. By leveraging geometric separation properties between intent clusters, we establish bounds on misclassification probability, which is the probabilistic guarantee $\hat{p}(y_t \mid z_t)$.

\begin{definition}[Distance to Centroid]
\label{def:centroid-dist}
    Let $\mathcal{Z} \subset \mathbb{R}^n$ be the latent embedding space induced by the multimodal encoder $P$, which jointly embeds textual and visual inputs $(I_t, U_t)$ along with their predicted intent labels $y_t$. Suppose $\mathcal{C} = \{c_1, c_2, \dots, c_K\}$ denotes the set of cluster centroids for $K$ intent classes, estimated from a calibration dataset $\mathcal{D}_c$ of size $N_c$. For a given input $(I_t, U_t)$ with latent embedding $z_t = P(I_t, U_t, y_t)$, define the Euclidean distance to the nearest centroid as:
    \[
    d_t = \min_{c_k \in \mathcal{C}} \| z_t - c_k \|_2 .
    \]
\end{definition}

\begin{definition}[Calibration Distribution]
\label{def:calib-dist}
    A calibration distribution, whose probability density function is $F_C: \mathbb{R}^+ \rightarrow [0,1]$, is the distribution of distances from correctly classified calibration samples to their corresponding class centroids. Given a new input, the function $F_C$ takes the distance $d_t$ to its nearest centroid $c^*$ as input, and returns a probability that a sample from another class lies beyond $d_t$:
    \begin{center}
        $\Pr\big[ || z_i - c^* ||_2 > d_t \,\big|\,  y_i \ne y_c^* \big]$, 
    \end{center}
    $y_c^*$ is the ground truth label.
    This distribution is approximated by using a labeled calibration dataset.
\end{definition}

\begin{theorem}[Latent Distance to Probabilistic Guarantee]
\label{thm:latent_distance}

Given a new input, we compute its distance $d_t$ to the nearest centroid and compute the probabilistic guarantee
\begin{equation}
    \hat{p}(y_t \mid z_t) = \big(1 - F_C(d_t)\big)\cdot \Pr\big[y_i \ne y_c^*\big] / \Pr\big[ || z_i - c^* ||_2 \le d_t \big].
\end{equation}
$\Pr\big[y_i \ne y_c^*\big]$ is the number of other-class samples over the total number of samples, and $\Pr\big[ || z_i - c^* ||_2 \le d_t \big]$ is the percentage of samples whose distance to $c^*$ is within $d_t$.
\end{theorem}

\begin{proof}
    According to Definition \ref{def:calib-dist}, given a new input whose latent space distance to its nearest centroid is $d_t$, then
    \begin{center}
        $F_C(d_t) = \Pr\big[ || z_i - c^* ||_2 > d_t \,\big|\,  y_i \ne y_c^* \big].$
    \end{center}
    Recall that the probabilistic guarantee $\hat{p}(y_t \mid z_t)$ is 
    \begin{center}
        $\Pr\big[ y_i = y_c^* \,\big|\, || z_i - c^* ||_2 \le d_t \big]$,
    \end{center}
    which is equal to
    \begin{center}
        $1 - \Pr\big[ y_i \ne y_c^* \,\big|\, || z_i - c^* ||_2 \le d_t \big]$,
    \end{center}
    where $y_i$ is the latent representation of a randomly selected sample.
    By the Bayesian rule, we know that
    \begin{align}
        \Pr \big[ y_i \ne y_c^* \,\big|\, || z_i - c^* ||_2 \le d_t \big]
        & = \frac{\Pr \big[ || z_i - c^* ||_2 \le d_t \,\big|\, y_i \ne y_c^* \big] \cdot  \Pr \big[ y_i \ne y_c^* \big]}{\Pr \big[ || z_i - c^* ||_2 \le d_t \big]} \\
        & = \frac{ (1 - \Pr \big[ || z_i - c^* ||_2 > d_t \,\big|\, y_i \ne y_c^* \big]) \cdot  \Pr \big[ y_i \ne y_c^* \big] }{ \Pr \big[ || z_i - c^* ||_2 \le d_t \big] } \\
        & = \frac{ (1 - F_C(d_t) ) \cdot  \Pr \big[ y_i \ne y_c^* \big] }{ \Pr \big[ || z_i - c^* ||_2 \le d_t \big] }.
    \end{align}
    $\Pr \big[ y_i \ne y_c^* \big] \text{ and } \Pr \big[ || z_i - c^* ||_2 \le d_t \big]$ are known constants obtained via the calibration dataset.

    Hence, we get the probabilistic guarantee
    \begin{align}
        \hat{p}(y_t \mid z_t) & = 1 - \Pr\big[ y_i \ne y_c^* \,\big|\, || z_i - c^* ||_2 \le d_t \big] \\
        & = 1 - \frac{ (1 - F_C(d_t) ) \cdot  \Pr \big[ y_i \ne y_c^* \big] }{ \Pr \big[ || z_i - c^* ||_2 \le d_t \big] }.
    \end{align}
    Proved.
\end{proof}

% \section{VLL Agent Online Interface}

% \label{app: interface}
% To show the practical capabilities of the VLL agent, we present a set of scenarios illustrating its interactive online interface. The system enables users to issue free-form natural language commands, receive executable plans, and iteratively refine objectives through proactive clarification queries. 

% Figure~\ref{fig: interface} shows three example scenarios. The agent can update airway risks for selected cities, re-route flights to reduce operational risks, and answer natural-language queries about specific airports. These examples demonstrate the system’s ability to combine conversational flexibility with reliable, real-time decision support.

% We present a video demonstration in \href{https://github.com/samuelLi05/airplane_simulator}{\textcolor{blue}{our GitHub repository}}.

% \begin{figure}[t]
%     \centering
%     \includegraphics[width=\textwidth]{figure/interfaces.pdf}
%     \caption{
%     Sample scenarios of the online interface for the VLL agent.
%     }
%     \label{fig: interface}
% \end{figure}

\end{document}